\documentclass[11pt,a4paper]{article}

\usepackage[hyperref]{emnlp-ijcnlp-2019}

\usepackage{amsfonts}
\usepackage{times}
\usepackage{latexsym}
\usepackage{amsmath}
\usepackage{url}
\usepackage{multirow}
\usepackage{graphicx}
\usepackage{graphics}
\usepackage{subfigure}
\usepackage{booktabs}
\usepackage{arydshln}

\newcommand{\hide}[1]{} 

\newcommand{\eg}{e.g.}

\aclfinalcopy 

\title{Learning Dynamic Context Augmentation for Global Entity Linking}

\author{Xiyuan Yang$^1$, Xiaotao Gu$^2$, Sheng Lin$^1$, Siliang Tang$^1$\thanks{~ Corresponding author.}, Yueting Zhuang$^1$, \\
    \textbf{Fei Wu$^1$, Zhigang Chen$^3$, Guoping Hu$^3$ \& Xiang Ren$^4$}\\
    $^1$Zhejiang University, $^2$University of Illinois at Urbana Champaign\\ 
    $^3$iFLYTEK Research, $^4$University of Southern California\\
    \texttt{\{yangxiyuan, shenglin, siliang, yzhuang, wufei\}@zju.edu.cn}, \\
    \texttt{xiaotao2@illinois.edu}, \texttt{\{zgchen, gphu\}@iflytek.com}, \\
    \texttt{xiangren@usc.edu}}
    
\date{}

\begin{document}
\maketitle
\setlength{\textfloatsep}{10pt plus 1.0pt minus 2.0pt}
\setlength{\intextsep}{10pt plus 1.0pt minus 2.0pt}
\setlength{\floatsep}{10pt plus 1.0pt minus 2.0pt}

\begin{abstract}
Despite of the recent success of collective entity linking (EL) methods, these ``global'' inference methods may yield sub-optimal results when the ``all-mention coherence" assumption breaks, and often suffer from high computational cost at the inference stage, due to the complex search space. In this paper, we propose a simple yet effective solution, called Dynamic Context Augmentation (DCA), for collective EL, which requires only one pass through the mentions in a document. DCA \textit{sequentially} accumulates context information to make efficient, collective inference, and can cope with different local EL models as a plug-and-enhance module.
We explore both supervised and reinforcement learning strategies for learning the DCA model. Extensive experiments\footnote{\scriptsize Code and data is available at https://github.com/YoungXiyuan/DCA} show the effectiveness of our model with different learning settings, base models, decision orders and attention mechanisms.
\end{abstract}

\section{Introduction}

\begin{figure}[h!]
\centering
\includegraphics[width=\columnwidth, angle=0]{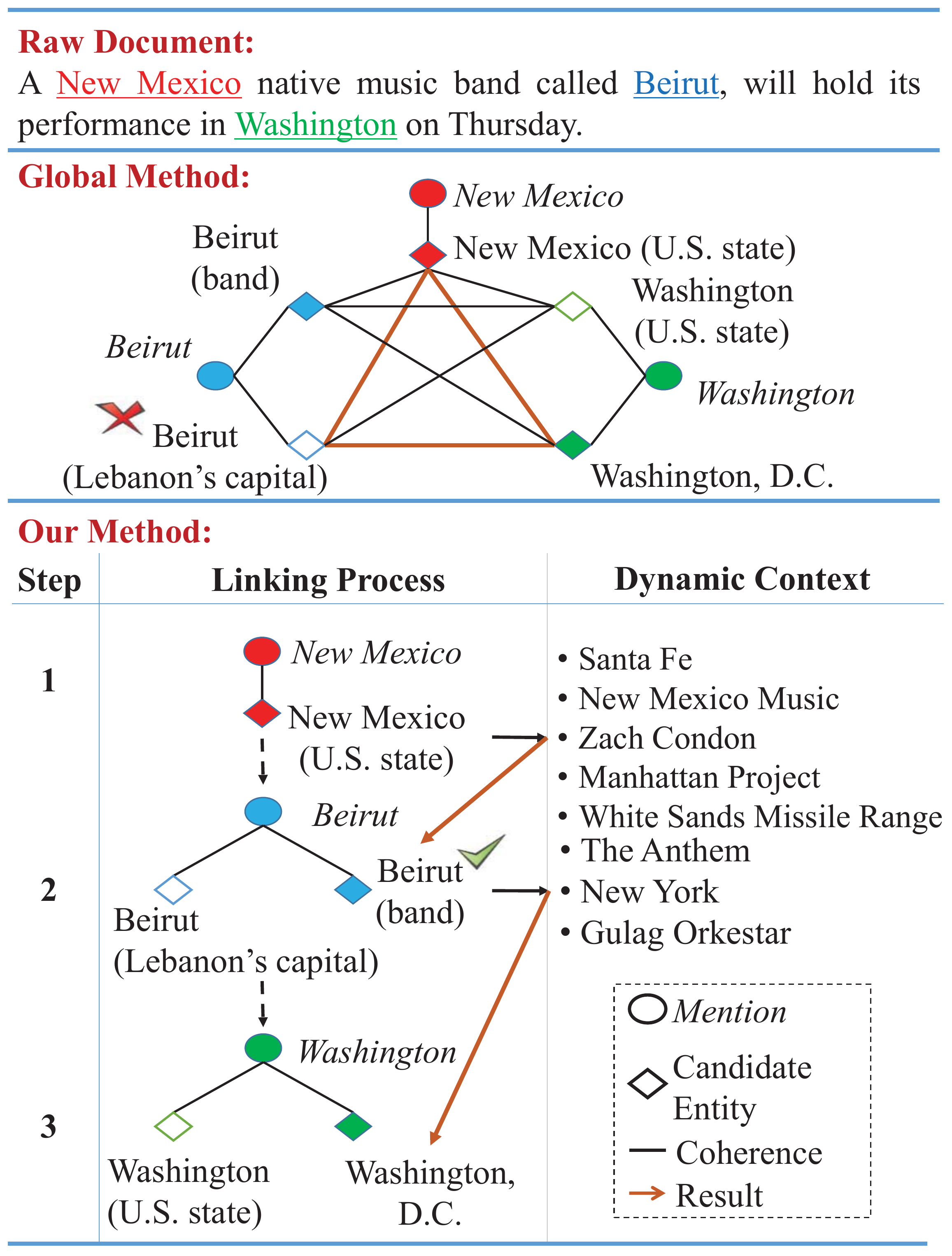}
\caption{\textbf{An Illustration of the Dynamic Context Augmentation process.} A traditional global EL model jointly optimizes the linking configuration after iterative calculations over all mentions, which is computationally expensive. In contrast, the DCA process only requires one pass of the document to accumulate knowledge from previously linked mentions to enhance fast future inference.
}

\label{fig:DCA}
\end{figure}

Linking mentions of entities in text to knowledge base entries (i.e., entity linking, or EL) is critical to understanding and structuring text corpora. In general, EL is approached by first obtaining candidate entities for each mention, and then identifying the true referent among the candidate entities. 
Prior distribution and local contexts, either in the form of hand-crafted features~\cite{ratinov2011local,shen2015entity} or dense embeddings~\cite{he2013learning,nguyen2016joint,francis2016capturing}, play key roles in distinguishing different candidates. However, in many cases, local features can be too sparse to provide sufficient information for disambiguation.

To alleviate this problem, various collective EL models have been proposed to globally optimize the inter-entity coherence between mentions in the same document
~\cite{hoffart2011robust,cheng2013relational,nguyen2014aida,alhelbawy2014graph,pershina2015personalized}.
Despite of their success, existing global EL models try to optimize the entire linking configuration of all mentions, with extra assumptions of either all-mention coherence or pairwise coherence \cite{phan2018pair}.
Such assumptions are against human intuitions, as they imply that no inference can be made until all mentions in a document have been observed.
Also, there usually exists a trade-off between accuracy and efficiency: state-of-the-art collective/global models suffer from high time complexity.
From the perspective of computational efficiency, optimal global configuration inference is NP-hard.
Approximation methods, such as loopy belief propagation \cite{ganea2017deep} or iterative substitutions \cite{shen2015entity}, are still computationally expensive due to the huge hypothesis space, and thus can hardly be scaled to handle large corpus.
Many previous works have discussed the urgent needs of more efficient linking system for production, both in time complexity~\cite{hughes2014trading} and memory consumption~\cite{blanco2015fast}.

In this paper, we propose a simple yet effective \textbf{D}ynamic \textbf{C}ontext \textbf{A}ugmentation (DCA) process to incorporate global signal for EL. 
As Figure \ref{fig:DCA} shows, in contrast to traditional global models, DCA only requires one pass through all mentions to achieve comparable linking accuracy.
The basic idea is to accumulate knowledge from previously linked entities as dynamic context to enhance later decisions.
Such knowledge come from not only the inherent properties (\eg, description, attributes) of previously linked entities, but also from their closely related entities, which empower the model with important associative abilities. 
In real scenarios, some previously linked entities may be irrelevant to the current mention. Some falsely linked entities may even introduce noise.
To alleviate error propagation, we further explore two strategies: (1) soft/hard attention mechanisms that favour the most relevant entities; (2) a reinforcement learning-based ranking model, which proves to be effective as reported in other information extraction tasks.

\smallskip
\noindent
\textbf{Contributions.} 
The DCA model forms a new linking strategy from the perspective of data augmentation and thus can serve as a plug-and-enhance module of existing linking models. The major contributions of this work are as follows:
(1) DCA can introduce topical coherence into local linking models without reshaping their original designs or structures; (2) Comparing to global EL models, DCA only requires one pass through all mentions, yielding better efficiency in both training and inference; (3) Extensive experiments show the effectiveness of our model under different learning settings, base models, decision orders and attention mechanisms.
\section{Background}

\subsection{Problem Definition}
Given a set of entity mentions $\mathcal{M} = \{m_1, ..., m_T\}$ in corpus $\mathcal{D}$, Entity Linking aims to link each mention $m_t$ to its corresponding gold entity $e_t^*$. Such a process is usually divided into two steps:
\emph{Candidate generation} first collects a set of possible (candidate) entities $\mathcal{E}_t = \{e_t^1, ..., e_t^{|\mathcal{E}_t|}\}$ for $m_t$;
\emph{Candidate ranking} is then applied to rank all candidates by likelihood. The linking system selects the top ranked candidate as the predicted entity $\hat{e}_t$.
The key challenge is to capture high-quality features of each entity mention for accurate entity prediction, especially when local contexts are too sparse to disambiguate all candidates.

We build our DCA model based on two existing local EL models. In this section, we first introduce the architecture of the base models, then present the proposed DCA model under the standard supervised learning framework. Since the DCA process can be naturally formed as a sequential decision problem, we also explore its effectiveness under the Reinforcement Learning framework. Detailed performance comparison and ablation studies are reported in Section \ref{sec:result}.
\subsection{Local Base Models for Entity Linking}
\label{sec:base}
We apply the DCA process in two popular local models with different styles: the first is a neural attention model named ETHZ-Attn~\cite{ganea2017deep}, the other is the Berkeley-CNN~\cite{francis2016capturing} model which is made up of multiple convolutional neural networks (CNN).

\medskip
\noindent\textbf{ETHZ-Attn.}
For each mention $m_t$ and a candidate $e^j_t \in \mathcal{E}_t$, three local features are considered: 
(1) \emph{Mention-entity Prior} $\hat{P}(e_t^j | m_t)$ is the empirical distribution estimated from massive corpus (\eg Wikipedia); 
(2) \emph{Context Similarity} $\Psi_C(m_t, e^j_t)$ measures the textual similarity between $e^j_t$ and the local context of $m_t$;
(3) \emph{Type Similarity} $\Psi_T(m_t, e^j_t)$ considers the similarity between the type of $e_t^j$ and contexts around $m_t$.
$\hat{P}(e_t^j | m_t)$ and $\Psi_C(m_t, e^j_t)$ are calculated in the same way as \cite{ganea2017deep}.
For $\Psi_T(m_t, e^j_t)$, we first train a typing system proposed by \cite{xu2018neural} on AIDA-train dataset, yielding 95\% accuracy on AIDA-A dataset. In the testing phase, the typing system predicts the probability distribution over all types (PER, GPE, ORG and UNK) for $m_t$, and outputs $\Psi_T(m_t, e^j_t)$ for each candidate accordingly. All local features are integrated by a two-layer feedforward neural network with 100 hidden units, as described in \cite{ganea2017deep}.

\medskip
\noindent\textbf{Berkeley-CNN.} The only difference between \textbf{ETHZ-Attn} and \textbf{Berkeley-CNN} is that, this model utilizes CNNs at different granularities to capture \emph{context similarity} $\Psi_C(m_t, e^j_t)$ between a mention's context and its target candidate entities.
\section{Dynamic Context Augmentation}
\label{sec:DCA}
\label{sec:dca}
\begin{figure}[]
\centering
\includegraphics[width=0.45\textwidth, angle=0]{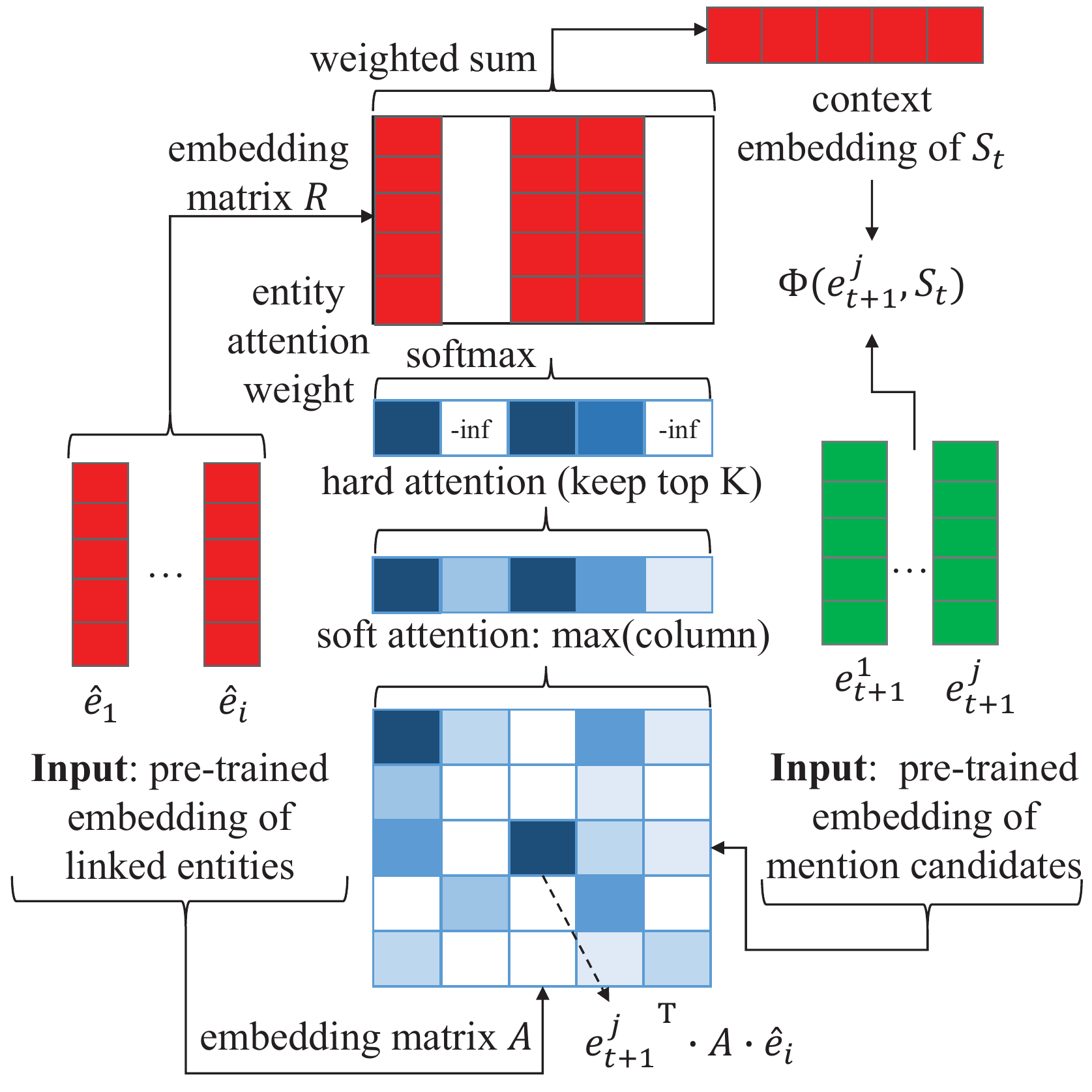}
\caption{Neural attention mechanism on the dynamic context. 
The soft attention module assigns higher weights to entities that are more relevant to the target mention.
The hard attention module only considers top $K$ entities as dynamic contexts.
}

\label{fig:attention}
\end{figure}

As Figure \ref{fig:DCA} demonstrates, the basic idea of DCA is to accumulate knowledge from previously linked entities as dynamic context to enhance later decisions.
Formally, denote the list of previously linked entities as $S_t = \{\hat{e}_1, ..., \hat{e}_t\}$, where each $\hat{e}_i$ is represented as an embedding vector.
The augmented context can be represented by accumulated features of all previous entities and their neighbors (e.g. by averaging their embeddings, in the simplest way).
In actual scenarios, some entities in $S_t$ are irrelevant, if not harmful, to the linking result of $m_{t+1}$. 
To highlight the importance of relevant entities while filtering noises, we also try to apply a neural attention mechanism on dynamic contexts (Figure \ref{fig:attention}).
For mention $m_{t+1}$, candidates that are more coherent with $S_t$ are preferred.
More specifically, we calculate the relevance score for each $\hat{e}_i \in S_t$ as
\begin{equation}
    u(\hat{e}_i) = \max_{e^j_{t+1} \in \mathcal{E}_{t+1}} {e_{t+1}^j}^\top \cdot A \cdot \hat{e}_i,
\end{equation}
where $A$ is a parameterized diagonal matrix. Top $K$ entities in $S_t$ are left to form dynamic context while the others are pruned. The relevance scores are transformed to attention weights with
\begin{equation}
    a(\hat{e}_i) = \frac{\exp[u(\hat{e}_i)]}{\sum_{\hat{e}_j \in S_t} \exp[u(\hat{e}_j)]}~.
\end{equation}
Thus, we can define a weighted coherence score between $e_{t+1}^j \in \mathcal{E}_{t+1}$ and $S_t$ as
\begin{equation}
    \Phi(e_{t+1}^j, S_t) = \sum_{\hat{e}_i \in S_t} a(\hat{e}_i) \cdot {e^j_{t+1}}^\top \cdot R \cdot \hat{e}_i,
\end{equation}
where $R$ is a learnable diagonal matrix. Such a coherence score will be later incorporated in the final representation of $e_{t+1}^j$.

To empower the linking model with associative ability, aside from previously linked entities, we also incorporate entities that are closely associated with entities in $S_t$. Specifically, for each $\hat{e}_i \in S_t$, we collect its neighborhood $\mathcal{N}(\hat{e}_i)$ consisting of Wikipedia entities that have inlinks pointing to $\hat{e}_i$. Denoting $S'_t$ as the union of $\{\mathcal{N}(\hat{e}_i) | \hat{e}_i \in S_t\}$, we define a similar weighted coherence score between $e_{t+1}^j \in \mathcal{E}_{t+1}$ and $S'_t$ as
\begin{equation}
    \Phi'(e_{t+1}^j, S'_t) = \sum_{\hat{e}_i \in S'_t} a'(\hat{e}_i) \cdot {e^j_{t+1}}^\top \cdot R' \cdot \hat{e}_i,
\end{equation}
where $a'$ is defined similarly to $a$, and $R'$ is a learnable diagonal matrix. 
The final representation $\vec{h_0}(m_{t+1}, e^j_{t+1})$ is the concatenation of $\Phi(e_{t+1}^j, S_t)$, $\Phi'(e_{t+1}^j, S'_t)$, $\Psi_T(m_t, e_{t+1}^j)$, $\Psi_C(m_t, e^j_{t+1})$ and $\log\hat{P}(e^j_{t+1} | m_{t+1})$. 
\section{Model Learning for DCA}
In this section, we explore different learning strategies for the linking model. Specifically, we present a Supervised Learning model, where the model is given all gold entities for training, and a Reinforcement Learning model, where the model explores possible linking results by itself in a long-term planning task.

\subsection{Supervised Ranking Method}
Given a mention-candidate pair $(m_t, e_t^j)$, the ranking model parameterized by $\theta$ accepts the feature vector $\vec{h_0}(m_{t}, e^j_{t})$ as input, and outputs the probability $P_\theta(e^j_t | m_t)$.
In this work, we use a two-layer feedforward neural network as the ranking model. We apply the max-margin loss as
\begin{equation*}
\begin{split}
    \mathcal{L_\theta} &= \sum_{D \in \mathcal{D}} \sum_{m_t \in D} \sum_{e_t^j \in \mathcal{E}_t} g_\theta(e_t^j, m_t),\\
    g_\theta(e_t^j, m_t) &= \max(0, \gamma - P_\theta(e^*_t | m_t) + P_\theta(e^j_t | m_t)).
\end{split} 
\end{equation*}
The learning process is to estimate the optimal parameter such that $\theta^* = \arg\min_\theta \mathcal{L}_\theta$.

Note that, in the Supervised Ranking model, dynamic contexts are provided by previous gold entities: $S_t = \{e_1^*, ..., e_t^*\}$, $S'_t = \bigcup\limits_{i=1}^{t} \mathcal{N}(e_i^*)$. In the testing phase, however, we do not have access to gold entities. Wrongly linked entities can introduce noisy contexts to future linking steps. To consider such long-term influences, we introduce an alternative Reinforcement Learning model in the next section.
\subsection{Reinforcement Learning Method}
Naturally, the usage of dynamic context augmentation forms a sequential decision problem, as each linking step depends on previous linking decisions.
Correct linking results provide valuable information for future decisions, while previous mistakes can lead to error accumulation. 
Reinforcement Learning (RL) algorithms have proven to be able to alleviate such accumulated noises in the decision sequence in many recent works \cite{narasimhan2016improving, feng2018relation}.
In this work, we propose an RL ranking model for DCA-enhanced entity linking.

\vspace{0.15cm}
\noindent\textbf{Agent}: The Agent is a candidate ranking model that has a similar architecture to~\cite{clark2016deep}, aiming to output the action preference $H_\theta(S_{t-1}, S'_{t-1}, A_t^j)$ of each linking action $A_t^j = (m_t \rightarrow e_t^j)$. It is a 2-layer feedforward neural network with following components:

\vspace{0.15cm}
\noindent{\bf Input Layer}: For each $( m_t, e_t^j )$ pair, DCA-RL extracts context-dependent features from $S_{t-1}, S'_{t-1}$, and concatenates them with other context-independent features to produce an $I$-dimensional input vector $\vec{h_0}(m_{t}, e^j_{t})$.

\vspace{0.15cm}
\noindent{\bf Hidden Layers}: Let $Drop(\vec x)$ be the dropout operation~\cite{srivastava2014dropout} and $ReLU(\vec x)$ be the rectifier nonlinearity \cite{nair2010rectified}. So the output $\vec {h}_1$ of the hidden layer is defined as:
\begin{equation}
\vec {h}_1 = Drop(ReLU(\vec{W}_1 \cdot \vec{h}_0 + \vec{b}_1)),
\end{equation}
where $\vec{W}_1$ is a $H_1 \times I$ weight matrix.

\vspace{0.15cm}
\noindent{\bf Output Layers}: This scoring layer is also fully connected layer of size 1.
\begin{equation}
\vec {h}_2 = \vec{W}_2 \cdot \vec{h}_1 + \vec{b}_2,
\end{equation}
where $\vec{W}_2$ is a $1 \times H_1$ weight matrix. In the end, all action preference would be normalized together using an exponential softmax distribution, getting their action probabilities $\pi_\theta(A_t^j|S_{t-1},S'_{t-1})$:

According to policy approximating methods, the best approximate policy may be stochastic. So we randomly sample the actions based on the softmax distribution during the training time, whereas deliberately select the actions with the highest ranking score at the test time.

\vspace{0.15cm}
\noindent
\textbf{Reward.}
The reward signals are quite sparse in our framework. For each trajectory, the Agent can only receive a reward signal after it finishes all the linking actions in a given document. Therefore the immediate reward of action $t$, $R_t = 0$, where $0 \leq t <T$, and $R_T = - (|\mathbb{M}_{e}|/T)$, where $T$ is total number of mentions in the source document, and $|\mathbb{M}_{e}|$ is the number of incorrectly linked mentions. Then the value $G_t$ (expected reward) of each previous state $S_t$ can be retraced back with a discount factor $\rho$ according to $R_T$:
\begin{equation}
G_t=\rho^{T-t} R_T
\end{equation}

To maximize the expected reward of all trajectories, the Agent utilizes the REINFORCE algorithm~\cite{sutton1998reinforcement} to compute Monte Carlo policy gradient over all trajectories, and perform gradient ascent on its parameters:
\begin{equation}
\theta \leftarrow \theta + \alpha \sum_t G_t \nabla_\theta \ln \pi_\theta(A_t^j|S_{t-1},S'_{t-1})
\end{equation}

In following sections, to fully investigate the effectiveness of the proposed method, we report and compare the performances of both the Supervised-learning model and the Reinforcement-learning model.
\section{Analysis of Computational Complexity}
\label{sec:complexity}
For each document $D$, the train and inference of the global EL models are heavily relied on the inter-entity coherence graph~$\Phi_g$. Many studies~\cite{ratinov2011local,globerson2016collective,yamada2016joint,ganea2017deep,le2018improving} obtain $\Phi_g$ by calculating all pairwise scores between two arbitrary elements ${e_i}^x$ and ${e_j}^y$ sampled independently from candidate sets $\mathcal{E}_i$ and $\mathcal{E}_j$ in the given document. It is obvious that $\Phi$ is intractable, and the computational complexity of $\Phi_g$ is
\begin{equation}
\mathcal{O}(\Phi_g)=\mathcal{O}(\sum_{i=1}^{T}\sum_{j=1,j\neq i}^{T}\sum_{{e_i}^x\in\mathcal{E}_i}^{|\mathcal{E}_i|}\sum_{{e_j}^y\in\mathcal{E}_j}^{|\mathcal{E}_j|}\Phi({e_i}^x,{e_j}^y))
\end{equation}, where $\Phi({e_i}^x,{e_j}^y)$ is a learnable score function. Thus, $\mathcal{O}(\Phi_g)$ is approximate to $\mathcal{O}(T^2 \times |\mathcal{E}|^2 \times \mathcal{I})$, where $|\mathcal{E}|$ is the average number of candidates per mention and $\mathcal{I}$ is the unit cost of pairwise function $\Phi$. In order to reduce $\mathcal{O}(\Phi_g)$, most previous models~\cite{hoffart2011robust,ganea2017deep,le2018improving,fang2019joint} have to hard prune their candidates into an extremely small size (e.g. $|\mathcal{E}|$=5). This will reduce the gold recall of candidate sets and also unsuitable for large scale production (e.g. entity disambiguation for dynamic web data like Twitter).

In contrast, the computational complexity of our model is $\mathcal{O}(T \times |\mathcal{E}| \times \mathcal{I} \times K )$, where $K$ is the key hyper-parameter described in Section \ref{sec:DCA} and is usually set to a small number. This indicates the response time of our method grow linearly as a function of $T \times |\mathcal{E}| $.

\begin{table}[t]
\centering
\small
\scalebox{1}{
\resizebox{\columnwidth}{!}{%
  \begin{tabular}{lcccc}
    \toprule[2pt]
    \textbf{Dataset} & \textbf{\begin{tabular}[x]{@{}c@{}} \#{} \\ mention\end{tabular}} &  \textbf{\begin{tabular}[x]{@{}c@{}} \#{} \\ doc\end{tabular}} & \textbf{\begin{tabular}[x]{@{}c@{}} Mentions\\per doc\end{tabular}} & \textbf{\begin{tabular}[x]{@{}c@{}} Gold\\recall\end{tabular}} \\
    \toprule[2pt]
    AIDA-train & 18448 & 946 & 19.5 & -\\
    AIDA-A & 4791 & 216 & 22.1 & 97.3 \\
    AIDA-B & 4485 & 231 & 19.4 & 98.3 \\
    \midrule
    MSNBC & 656 & 20 & 32.8 & 98.5\\
    AQUAINT & 727 & 50 & 14.5 & 94.2 \\
    ACE2004 & 257 & 36 & 7.1 & 90.6 \\
    CWEB & 11154 & 320 & 34.8 & 91.1 \\
    WIKI & 6821 & 320 & 21.3 & 92.4\\
    \bottomrule[2pt]
\end{tabular}%
}
}
\caption{\textbf{Dataset Statistics.} \emph{Gold recall} is the percentage of mentions for which the candidate entities contain the ground truth entity. 
}
\label{tab:dataset}
\end{table}

\section{Experiment}
\label{sec:result}

\subsection{Experiment Setup}
\label{sec:setup}
\textbf{Datasets.} Following our predecessors, we train and test all models on the public and widely used AIDA CoNLL-YAGO dataset \cite{hoffart2011robust}. The target knowledge base is Wikipedia. The corpus consists of 946 documents for training, 216 documents for development and 231 documents for testing (AIDA-train/A/B respectively). To evaluate the generalization ability of each model, we apply cross-domain experiments following the same setting in~\cite{ganea2017deep, le2018improving, yang2018collective}. Models are trained on AIDA-train, and evaluated on five popular public datasets: AQUAINT~\cite{milne2008learning}, MSNBC~\cite{cucerzan2007large}, ACE2004~\cite{ratinov2011local}, CWEB~\cite{guo2016robust} and WIKI \cite{guo2016robust}. The statistics of these datasets are available in Table~\ref{tab:dataset}. In the candidate generation step, we directly use the candidates provided by the Ment-Norm system \cite{le2018improving}\footnote{https://github.com/lephong/mulrel-nel}, and their quality is also listed in Table \ref{tab:dataset}.\\

\begin{table}[t]
\centering
\resizebox{\columnwidth}{!}{%
\begin{tabular}{lc}
  \toprule[2pt]
  \textbf{System}  & \textbf{\begin{tabular}[x]{@{}c@{}} In-KB acc. (\%)\end{tabular}} \\
  \toprule[2pt]
  AIDA-light~\cite{nguyen2014aida} & 84.8 \\
  WNED~\cite{guo2016robust} & 89.0 \\
  Global-RNN~\cite{nguyen2016joint} & 90.7  \\
  MulFocal-Att~\cite{globerson2016collective} & 91.0 \\
  Deep-ED~\cite{ganea2017deep} & 92.22\\
  Ment-Norm~\cite{le2018improving} & 93.07 \\
  \midrule
  Prior ($p(e|m)$)~\cite{ganea2017deep} &  71.51 \\
  \midrule
  \emph{Berkeley-CNN (Sec.~\ref{sec:base})}  & 84.21 \\
  \emph{Berkeley-CNN~+~DCA-SL} & 92.72 $\pm$ 0.3\\
  \emph{Berkeley-CNN~+~DCA-RL} & 92.37 $\pm$ 0.1 \\
  \emph{ETHZ-Attn (Sec.~\ref{sec:base})} & 90.88 \\
  \emph{ETHZ-Attn~+~DCA-SL} &  \textbf{{94.64}} $\pm$ 0.2 \\
  \emph{ETHZ-Attn~+~DCA-RL} & {93.73} $\pm$ 0.2 \\
  \bottomrule[2pt]
\end{tabular}%
}
\caption{\textbf{In-domain Performance Comparison on the AIDA-B Dataset.} For our method we show 95\% confidence intervals obtained over 5 runs. DCA-based models achieve the best reported scores on this benchmark.
}
\label{tab:conll}
\end{table}

\begin{table*}[t]
\centering
\scalebox{1}{
\resizebox{\textwidth}{!}{%
\begin{tabular}{lccccc}
  \toprule[2pt]
  \textbf{System} & \textbf{MSBNC} & \textbf{AQUAINT} & \textbf{ACE2004} & \textbf{CWEB} &\textbf{WIKI} \\
  \toprule[2pt]
  AIDA~\cite{hoffart2011robust} & 79 & 56 & 80 & 58.6 & 63 \\
  GLOW~\cite{ratinov2011local} & 75 & 83 & 82 & 56.2 & 67.2 \\
  RI~\cite{cheng2013relational} & 90 & \textbf{90} & 86 & 67.5 & 73.4 \\ 
  WNED~\cite{guo2016robust} & 92 & 87 & 88 & 77 & \textbf{84.5} \\
  Deep-ED~\cite{ganea2017deep} &  93.7 & 88.5 & 88.5 & \textbf{77.9} & 77.5 \\
  Ment-Norm~\cite{le2018improving} & 93.9 & 88.3 & 89.9 & 77.5 & 78.0 \\
  \midrule
  Prior ($p(e|m)$)~\cite{ganea2017deep} & 89.3 & 83.2 & 84.4 & 69.8 & 64.2 \\
  \midrule
  \emph{Berkeley-CNN (Section \ref{sec:base})} & 89.05 & 80.55 & 87.32 & 67.97 & 60.27 \\
  \emph{Berkeley-CNN~+~DCA-SL} & 93.38 $\pm$ 0.2 & 85.63 $\pm$ 0.3 & 88.73 $\pm$ 0.3 & 71.01 $\pm$ 0.1 & 72.55  $\pm$ 0.2 \\
  \emph{Berkeley-CNN~+~DCA-RL} & 93.65 $\pm$ 0.2 & 88.53 $\pm$ 0.3 & 89.73 $\pm$ 0.4 & 72.66 $\pm$ 0.4 & 73.98 $\pm$ 0.2 \\
  \emph{ETHZ-Attn (Section  \ref{sec:base})} & 91.97 & 84.06 & 86.92 & 70.07 & 74.37  \\
  \emph{ETHZ-Attn~+~DCA-SL} &  \textbf{94.57} $\pm$ 0.2 & 87.38 $\pm$ 0.5 & 89.44 $\pm$ 0.4 & 73.47 $\pm$ 0.1 & 78.16 $\pm$ 0.1 \\
  \emph{ETHZ-Attn~+~DCA-RL} & 93.80 $\pm$ 0.0 & 88.25 $\pm$ 0.4 & \textbf{90.14} $\pm$ 0.0 & 75.59 $\pm$ 0.3 & 78.84 $\pm$ 0.2 \\
  \bottomrule[2pt]
\end{tabular}%
}
}
\caption{\textbf{Performance Comparison on Cross-domain Datasets using F1 score (\%)}. The best results are in bold. Note that our own results all retain two decimal places. Other results with uncertain amount of decimal places are directly retrieved from their original paper.}
\label{tab:soa-cross}
\end{table*}

\medskip
\noindent
\textbf{Compared Methods.}
We compare our methods with following existing systems that report state-of-the-art results on the test datasets: \textbf{AIDA-light}~\cite{nguyen2014aida} uses a kind of two-stage collective mapping algorithm and designs several domain or category related coherence features.
\textbf{WNED}~\cite{guo2016robust} applies random walks on carefully built disambiguation graphs and uses a greedy, iterative and global disambiguation algorithm based on Information Theory.
\textbf{Global-RNN}~\cite{nguyen2016joint} develops a framework based on convolutional neural networks and recurrent neural networks to simultaneously model the local and global features.
\textbf{MulFocal-Att}~\cite{globerson2016collective} adopts a coherence model with a multi-focal attention mechanism.
\textbf{Deep-ED}~\cite{ganea2017deep} leverages learned neural representations, and uses a deep learning model combined with a neural attention mechanism and graphical models.
\textbf{Ment-Norm}~\cite{le2018improving} improving the \textbf{Deep-ED} model by modeling latent relations between mentions.

For a fair comparison with prior work, we use the same input as the \textbf{WNED}, \textbf{Deep-ED} and \textbf{Ment-Norm} (models proposed after 2016), and report the performance of our model with both Supervised Learning (\textbf{DCA-SL}) and Reinforcement Learning (\textbf{DCA-RL}). We won't compare our models with the \textbf{RLEL}~\cite{fang2019joint} which is a deep reinforcement learning based LSTM model. There are two reasons: 1) \textbf{RLEL} uses optimized candidate sets with smaller candidate size and higher gold recall than ours and the listed baselines. 2) \textbf{RLEL} uses additional training set from Wikipedia data. \cite{fang2019joint} doesn't release either their candidate sets or updated training corpus, so the comparison with their work would be unfair for us.

\medskip
\noindent\textbf{Hyper-parameter Setting.} We coarsely tune the hyper-parameters according to model performance on AIDA-A. We set the dimensions of word embedding and entity embedding to 300, where the word embedding and entity embedding are publicly released by \cite{pennington2014glove} and  \cite{ganea2017deep} respectively. Hyper-parameters of the best validated model are: $K=7$, $I=5$, $H_1=100$, and the probability of dropout is set to 0.2. Besides, the rank margin $\gamma=0.01$ and the discount factor $\rho=0.9$. We also regularize the \emph{Agent} model as adopted in \cite{ganea2017deep} by constraining the sum of squares of all weights in the linear layer with $MaxNorm=4$. When training the model,  we use Adam~\cite{kingma2014adam} with learning rate of 2e-4 until validation accuracy exceeds 92.8\%, afterwards setting it to 5e-5. \\

\begin{table}[]
\centering
\scalebox{0.88}{
\resizebox{\columnwidth}{!}{%
    \begin{tabular}{c|cc}
    \toprule[2pt]
    \multirow{2}{*}{\textbf{System}} & \multicolumn{2}{c}{\textbf{In-KB acc. (\%)}} \\ \cline{2-3} 
                                     & \textbf{SL}           & \textbf{RL}           \\ \hline
    ETHZ-Attn (Section \ref{sec:base})                   & 90.88                 & -                     \\
    ETHZ-Attn + 1-hop DCA           & 93.69                 & 93.20                 \\
    ETHZ-Attn + 2-hop DCA           & 94.47                 & 93.76                 \\ 
    \bottomrule[2pt] 
    \end{tabular}%
}
}
\caption{\textbf{Ablation Study on Neighbor Entities.} We compare the performance of DCA with or without neighbor entities (i.e., 2-hop vs. 1-hop).}
\label{tab:abl_hop}
\end{table}

\begin{figure*}[t]
\centering
\includegraphics[width=0.95\textwidth, angle=0]{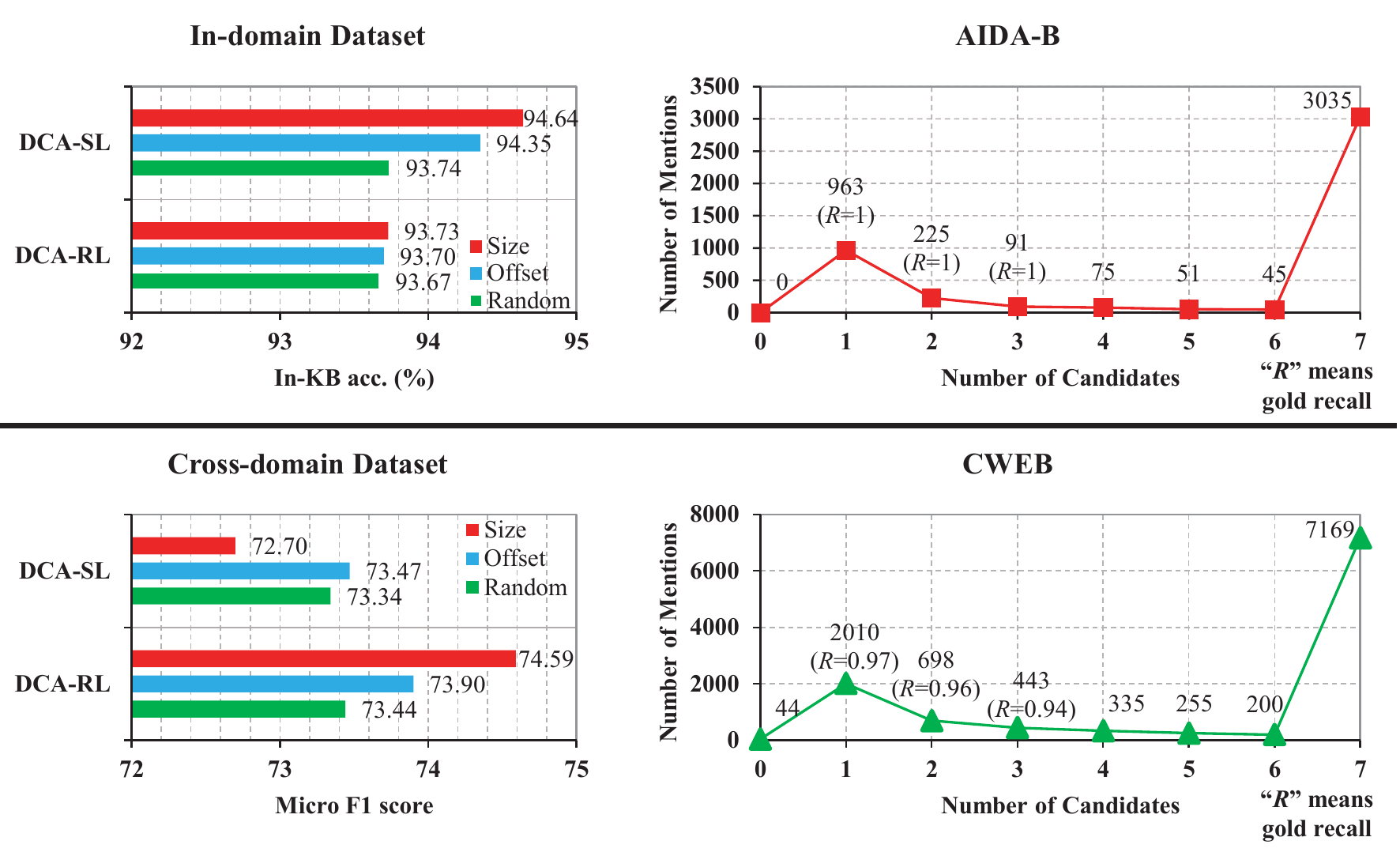}
\caption{\textbf{Ablation Study on Different Decision Orders.} We test on both in-domain (AIDA-B) and cross-domain (CWEB) datasets, using ETHZ-Attn as the local model.}
\label{fig:abl_order}
\end{figure*}

\vspace{-0.4cm}
\subsection{Overall Performance Comparison}
Starting with an overview of the end-task performance, we compare DCA (using SL or RL) with several state-of-the-art systems on in-domain and cross-domain datasets. We follow prior work and report in-KB accuracy for AIDA-B and micro F1 scores for the other test sets.

Table \ref{tab:conll} summarizes results on the AIDA-B dataset, and shows that DCA-based models achieve the highest in-KB accuracy and outperforms the previous state-of-the-art neural system by near 1.6\% absolute accuracy. Moreover, compared with the base models, dynamic context augmentation significantly improve absolute in-KB accuracy in models Berkeley-CNN (more than 8\%) and ETHZ-Attn (3.3\% on average). Note that, our DCA model outperforms existing global models with the same local model (\textbf{Global-RNN} uses Berkeley-CNN as base model, \textbf{Deep-ED} and \textbf{Ment-Norm} use ETHZ-Attn as the local model).

Table \ref{tab:soa-cross} shows the results on the five cross-domain datasets. As shown, none of existing methods can consistently win on all datasets. DCA-based models achieve state-of-the-art performance on the MSBNC and the ACE2004 dataset. On remaining datasets, DCA-RL achieves comparable performance with other complex global models. In addition, RL-based models show on average 1.1\% improvement on F1 score over the SL-based models across all the cross-domain datasets. At the same time, DCA-based methods are much more efficient, both in time complexity and in resource requirement. Detailed efficiency analysis will be presented in following sections.

\begin{figure*}[h]
\centering
\includegraphics[width=1.0\textwidth, angle=0]{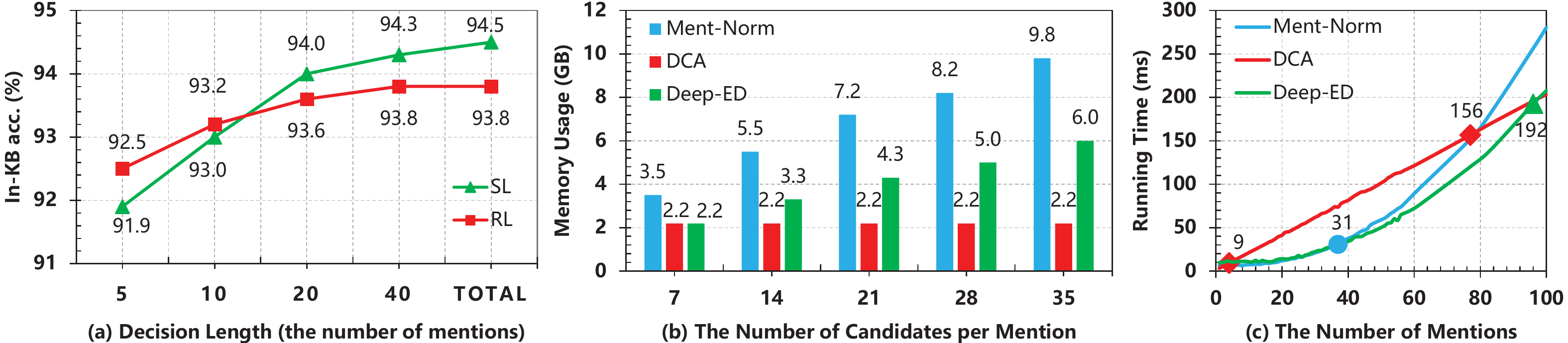}
\caption{(a) \textbf{In-KB accuracy} as a function of decision length on AIDA-B dataset; (b) \textbf{Memory usage} as a function of the number of candidates on AIDA-B dataset; (c) \textbf{Runtime cost} (at inference time) as a function of the number of mentions. (each dot represents a document of AIDA-B with $|\mathcal{E}|=35$).}
\label{fig:abl_total}
\end{figure*}

\begin{table}[]
\centering
\scalebox{0.9}{
\resizebox{\columnwidth}{!}{%
\begin{tabular}{c|cc}
\toprule[2pt]
\multirow{2}{*}{\textbf{System}} & \multicolumn{2}{c}{\textbf{In-KB acc. (\%)}} \\ \cline{2-3} 
                                 & \textbf{SL}           & \textbf{RL}           \\ \hline
DCA + Average Sum                & 94.05                 & 93.56                 \\
DCA + Soft Attention             & 94.23                 & 93.69                 \\
DCA + Soft\&Hard Attention     & 94.47                 & 93.76                 \\ \bottomrule[2pt] 
\end{tabular}%
}
}
\caption{\textbf{Study on Different Attention Mechanisms on the Dynamic Context.} Results (using ETHZ-Attn as local model) show that different attention mechanisms have similar impact on the performance.}
\label{tab:abl_attention}
\end{table}

\subsection{Performance Analysis}
\noindent
\textbf{1. Impact of decision order.}
As the DCA model consecutively links and adds all the mentions in a document, the linking order may play a key role in the final performance. In this work, we try three different linking orders: \emph{Offset} links all mentions by their natural orders in the original document; \emph{Size} first links mentions with smaller candidate sizes, as they tend to be easier to link; The baseline method is to link all mentions in a \emph{Random} order.
Figure \ref{fig:abl_order} shows the performance comparison on the AIDA-B and the CWEB dataset. As shown, in general, \emph{Size} usually leads to better performance than \emph{Offset} and \emph{Random}. However, the DCA-SL model shows poor performance on the CWEB dataset with \emph{Size} order. This is mainly because the CWEB dataset is automatically generated rather than curated by human, and thus contains many noisy mentions. Some mentions in CWEB with less than three candidates are actually bad cases , where none of the candidates is the actual gold entity. Thus, such mentions will always introduce wrong information to the model, which leads to a worse performance. In contrast, the AIDA-B dataset does not have such situations. The DCA-RL model, however, still has strong performance on the CWEB dataset, which highlights its robustness to potential noises.\\

\noindent
\textbf{2. Effect of neighbor entities.}
In contrast to traditional global models, we include both previously linked entities and their close neighbors for global signal. Table \ref{tab:abl_hop} shows the effectiveness of this strategy. We observe that incorporating these neighbor entities (2-hop) significantly improve the performance (compared to 1-hop) by introducing more related information. And our analysis shows that on average 0.72\% and 3.56\% relative improvement of 2-hop DCA-(SL/RL) over 1-hop DCA-(SL/RL) or baseline-SL (without DCA) is statistically significant (with P-value $<$ 0.005). This is consistent with our design of DCA.\\

\noindent
\textbf{3. Study of different attention mechanisms.}
Table \ref{tab:abl_attention} shows the performance comparison by replacing the attention module described in Section \ref{sec:DCA} with different variants. \emph{Average Sum} treats all previously linked entities equally with a uniform distribution. \emph{Soft Attention} skips the pruning step for entities with low weight scores. \emph{Soft\&Hard Attention} stands for the strategy used in our model. It is obvious that the attention mechanism does show positive influence on the linking performance compared with \emph{Average Sum}. Hard pruning brings slight further improvement.\\

\noindent
\textbf{4. Impact of decision length.}
As wrongly linked entities can introduce noise to the model, there exists a trade-off in DCA: involving more previous entities (longer historical trajectory) provides more information, and also more noise. Figure (\ref{fig:abl_total}.a) shows how the performance of DCA changes with the number of previous entities involved. We observe that longer historical trajectories usually have a positive influence on the performance of DCA. The reason is that our attention mechanism could effectively assess and select relevant contexts for each entity mention on the fly, thus reducing potential noise.

\smallskip
\noindent
\subsection{Analysis on Time Complexity}
As discussed in Sec.~\ref{sec:complexity}, the running time of a DCA enhanced model may rise linearly when the average number of candidates per mention (i.e., $|\mathcal{E}|$) increases, while the global EL model increases exponentially. To validate the theory we empirically investigate the scalability of DCA, and carefully select two global EL models Ment-Norm~\cite{le2018improving} and Deep-ED~\cite{ganea2017deep} as our baselines. The reason for this choice is that our final model shares the same local model as their models, which excludes other confounding factors like implementation details. As Figure (\ref{fig:abl_total}.c) shows, when  $|\mathcal{E}|$  increases, the running time of these two global EL models increases shapely, while our DCA model grows linearly. On the other hand, we also observed that the resources required by the DCA model are insensitive to $|\mathcal{E}|$. For example, as shown in Figure (\ref{fig:abl_total}.b), the memory usage of Ment-Norm and Deep-ED significantly rises as more candidates are considered, while the DCA model remains a relatively low memory usage all the time. We also measure the power consumption of Ment-Norm and DCA models, and we find that the DCA model saves up to 80\% of the energy consumption over the Ment-Norm, which is another advantage for large scale production.

\section{Related Work}
\label{sec:related_work}
Local EL methods disambiguate each mention independently according to their local contexts \cite{yamada2016joint, Chen2017Bilinear, globerson2016collective, raiman2018deeptype}. 
The performance is limited when sparse local contexts fail to provide sufficient disambiguation evidence.

To alleviate this problem, global EL models jointly optimize the entire linking configuration.
The key idea is to maximize a global coherence/similarity score between all linked entities \cite{hoffart2011robust,ratinov2011local,cheng2013relational,nguyen2014aida,alhelbawy2014graph,pershina2015personalized,guo2016robust,globerson2016collective,ganea2017deep,le2018improving,yang2018collective,fang2019joint,xue2019neural}.
Despite of its significant improvement in accuracy, such global methods suffer from high complexity.
To this end, some works try to relax the assumption of all-mention coherence, e.g. with pairwise coherence, to improve efficiency \cite{phan2018pair}, but exact inference remains an NP-hard problem.
Approximation methods are hence proposed to achieve reasonably good results with less cost.
\cite{shen2012linden} propose the iterative substitution method to greedily substitute linking assignment of one mention at a time that can improve the global objective.
Another common practice is to use Loopy Belief Propagation for inference \cite{ganea2017deep,le2018improving}.
Both approximation methods iteratively improve the global assignment, but are still computationally expensive with unbounded number of iterations.
In contrast, the proposed DCA method only requires one pass through the document. Global signals are accumulated as dynamic contexts for local decisions, which significantly reduces computational complexity and memory consumption.
\section{Conclusions}
In this paper we propose Dynamic Context Augmentation as a plug-and-enhance module for local Entity Linking models. In contrast to existing global EL models, DCA only requires one pass through the document. 
To incorporate global disambiguation signals, DCA accumulates knowledge from previously linked entities for fast inference.
Extensive experiments on several public benchmarks with different learning settings, base models, decision orders and attention mechanisms demonstrate both the effectiveness and efficiency of DCA-based models. The scalability of DCA-based models make it possible to handle large-scale data with long documents.
Related code and data has been published and may hopefully benefit the community.
\section*{Acknowledgments}

This work has been supported in part by National Key Research and Development Program of China (SQ2018AAA010010), NSFC (No.61751209, U1611461), Zhejiang University-iFLYTEK Joint Research Center, Zhejiang University-Tongdun Technology Joint Laboratory of Artificial Intelligence, Hikvision-Zhejiang University Joint Research Center, Chinese Knowledge Center of Engineering Science and Technology (CKCEST), Engineering Research Center of Digital Library, Ministry of Education. Xiang Ren's research has been supported in part by National Science Foundation SMA 18-29268, DARPA MCS and GAILA, IARPA BETTER, Schmidt Family Foundation, Amazon Faculty Award, Google Research Award, Snapchat Gift and JP Morgan AI Research Award. Finally, we would like to thank Qi Zhang from Sogou, Inc. and all the collaborators in INK research lab for their constructive feedback on this work.

\bibliography{emnlp-ijcnlp-2019}

\begin{thebibliography}{35}
\expandafter\ifx\csname natexlab\endcsname\relax\def\natexlab#1{#1}\fi

\bibitem[{Alhelbawy and Gaizauskas(2014)}]{alhelbawy2014graph}
Ayman Alhelbawy and Robert Gaizauskas. 2014.
\newblock Graph ranking for collective named entity disambiguation.
\newblock In \emph{Proceedings of the 52nd Annual Meeting of the Association
  for Computational Linguistics (Volume 2: Short Papers)}, volume~2, pages
  75--80.

\bibitem[{Blanco et~al.(2015)Blanco, Ottaviano, and Meij}]{blanco2015fast}
Roi Blanco, Giuseppe Ottaviano, and Edgar Meij. 2015.
\newblock Fast and space-efficient entity linking for queries.
\newblock In \emph{Proceedings of the Eighth ACM International Conference on
  Web Search and Data Mining}, pages 179--188. ACM.

\bibitem[{Chen et~al.(2017)Chen, Wei, Liu, Li, Yu, and Zhu}]{Chen2017Bilinear}
Hui Chen, Baogang Wei, Yonghuai Liu, Yiming Li, Jifang Yu, and Wenhao Zhu.
  2017.
\newblock Bilinear joint learning of word and entity embeddings for entity
  linking.
\newblock \emph{Neurocomputing}.

\bibitem[{Cheng and Roth(2013)}]{cheng2013relational}
Xiao Cheng and Dan Roth. 2013.
\newblock Relational inference for wikification.
\newblock In \emph{Proceedings of the 2013 Conference on Empirical Methods in
  Natural Language Processing}, pages 1787--1796.

\bibitem[{Clark and Manning(2016)}]{clark2016deep}
Kevin Clark and Christopher~D Manning. 2016.
\newblock Deep reinforcement learning for mention-ranking coreference models.
\newblock In \emph{Proceedings of the 2016 Conference on Empirical Methods in
  Natural Language Processing}, pages 2256--2262.

\bibitem[{Cucerzan(2007)}]{cucerzan2007large}
Silviu Cucerzan. 2007.
\newblock Large-scale named entity disambiguation based on wikipedia data.
\newblock In \emph{Proceedings of the 2007 Joint Conference on Empirical
  Methods in Natural Language Processing and Computational Natural Language
  Learning (EMNLP-CoNLL)}.

\bibitem[{Fang et~al.(2019)Fang, Cao, Li, Zhang, Zhang, and
  Liu}]{fang2019joint}
Zheng Fang, Yanan Cao, Qian Li, Dongjie Zhang, Zhenyu Zhang, and Yanbing Liu.
  2019.
\newblock Joint entity linking with deep reinforcement learning.
\newblock In \emph{The World Wide Web Conference}, pages 438--447. ACM.

\bibitem[{Feng et~al.(2018)Feng, Huang, Zhang, Yang, and
  Zhu}]{feng2018relation}
Jun Feng, Minlie Huang, Yijie Zhang, Yang Yang, and Xiaoyan Zhu. 2018.
\newblock Relation mention extraction from noisy data with hierarchical
  reinforcement learning.
\newblock \emph{arXiv preprint arXiv:1811.01237}.

\bibitem[{Francis-Landau et~al.(2016)Francis-Landau, Durrett, and
  Klein}]{francis2016capturing}
Matthew Francis-Landau, Greg Durrett, and Dan Klein. 2016.
\newblock Capturing semantic similarity for entity linking with convolutional
  neural networks.
\newblock \emph{arXiv preprint arXiv:1604.00734}.

\bibitem[{Ganea and Hofmann(2017)}]{ganea2017deep}
Octavian-Eugen Ganea and Thomas Hofmann. 2017.
\newblock Deep joint entity disambiguation with local neural attention.
\newblock \emph{arXiv preprint arXiv:1704.04920}.

\bibitem[{Globerson et~al.(2016)Globerson, Lazic, Chakrabarti, Subramanya,
  Ringaard, and Pereira}]{globerson2016collective}
Amir Globerson, Nevena Lazic, Soumen Chakrabarti, Amarnag Subramanya, Michael
  Ringaard, and Fernando Pereira. 2016.
\newblock Collective entity resolution with multi-focal attention.
\newblock In \emph{Proceedings of the 54th Annual Meeting of the Association
  for Computational Linguistics (Volume 1: Long Papers)}, volume~1, pages
  621--631.

\bibitem[{Guo and Barbosa(2016)}]{guo2016robust}
Zhaochen Guo and Denilson Barbosa. 2016.
\newblock Robust named entity disambiguation with random walks.
\newblock \emph{Semantic Web}, (Preprint):1--21.

\bibitem[{He et~al.(2013)He, Liu, Li, Zhou, Zhang, and Wang}]{he2013learning}
Zhengyan He, Shujie Liu, Mu~Li, Ming Zhou, Longkai Zhang, and Houfeng Wang.
  2013.
\newblock Learning entity representation for entity disambiguation.
\newblock In \emph{Proceedings of the 51st Annual Meeting of the Association
  for Computational Linguistics (Volume 2: Short Papers)}, volume~2, pages
  30--34.

\bibitem[{Hoffart et~al.(2011)Hoffart, Yosef, Bordino, F{\"u}rstenau, Pinkal,
  Spaniol, Taneva, Thater, and Weikum}]{hoffart2011robust}
Johannes Hoffart, Mohamed~Amir Yosef, Ilaria Bordino, Hagen F{\"u}rstenau,
  Manfred Pinkal, Marc Spaniol, Bilyana Taneva, Stefan Thater, and Gerhard
  Weikum. 2011.
\newblock Robust disambiguation of named entities in text.
\newblock In \emph{Proceedings of the Conference on Empirical Methods in
  Natural Language Processing}, pages 782--792. Association for Computational
  Linguistics.

\bibitem[{Hughes et~al.(2014)Hughes, Nothman, and Curran}]{hughes2014trading}
Kristy Hughes, Joel Nothman, and James~R Curran. 2014.
\newblock Trading accuracy for faster named entity linking.
\newblock In \emph{Proceedings of the Australasian Language Technology
  Association Workshop 2014}, pages 32--40.

\bibitem[{Kingma and Ba(2014)}]{kingma2014adam}
Diederik~P Kingma and Jimmy Ba. 2014.
\newblock Adam: A method for stochastic optimization.
\newblock \emph{arXiv preprint arXiv:1412.6980}.

\bibitem[{Le and Titov(2018)}]{le2018improving}
Phong Le and Ivan Titov. 2018.
\newblock Improving entity linking by modeling latent relations between
  mentions.
\newblock \emph{arXiv preprint arXiv:1804.10637}.

\bibitem[{Milne and Witten(2008)}]{milne2008learning}
David Milne and Ian~H Witten. 2008.
\newblock Learning to link with wikipedia.
\newblock In \emph{Proceedings of the 17th ACM conference on Information and
  knowledge management}, pages 509--518. ACM.

\bibitem[{Nair and Hinton(2010)}]{nair2010rectified}
Vinod Nair and Geoffrey~E Hinton. 2010.
\newblock Rectified linear units improve restricted boltzmann machines.
\newblock In \emph{Proceedings of the 27th international conference on machine
  learning (ICML-10)}, pages 807--814.

\bibitem[{Narasimhan et~al.(2016)Narasimhan, Yala, and
  Barzilay}]{narasimhan2016improving}
Karthik Narasimhan, Adam Yala, and Regina Barzilay. 2016.
\newblock Improving information extraction by acquiring external evidence with
  reinforcement learning.
\newblock \emph{arXiv preprint arXiv:1603.07954}.

\bibitem[{Nguyen et~al.(2014)Nguyen, Hoffart, Theobald, and
  Weikum}]{nguyen2014aida}
Dat~Ba Nguyen, Johannes Hoffart, Martin Theobald, and Gerhard Weikum. 2014.
\newblock Aida-light: High-throughput named-entity disambiguation.
\newblock \emph{LDOW}, 1184.

\bibitem[{Nguyen et~al.(2016)Nguyen, Fauceglia, Muro, Hassanzadeh, Gliozzo, and
  Sadoghi}]{nguyen2016joint}
Thien~Huu Nguyen, Nicolas Fauceglia, Mariano~Rodriguez Muro, Oktie Hassanzadeh,
  Alfio~Massimiliano Gliozzo, and Mohammad Sadoghi. 2016.
\newblock Joint learning of local and global features for entity linking via
  neural networks.
\newblock In \emph{Proceedings of COLING 2016, the 26th International
  Conference on Computational Linguistics: Technical Papers}, pages 2310--2320.

\bibitem[{Pennington et~al.(2014)Pennington, Socher, and
  Manning}]{pennington2014glove}
Jeffrey Pennington, Richard Socher, and Christopher Manning. 2014.
\newblock Glove: Global vectors for word representation.
\newblock In \emph{Proceedings of the 2014 conference on empirical methods in
  natural language processing (EMNLP)}, pages 1532--1543.

\bibitem[{Pershina et~al.(2015)Pershina, He, and
  Grishman}]{pershina2015personalized}
Maria Pershina, Yifan He, and Ralph Grishman. 2015.
\newblock Personalized page rank for named entity disambiguation.
\newblock In \emph{Proceedings of the 2015 Conference of the North American
  Chapter of the Association for Computational Linguistics: Human Language
  Technologies}, pages 238--243.

\bibitem[{Phan et~al.(2018)Phan, Sun, Tay, Han, and Li}]{phan2018pair}
Minh~C Phan, Aixin Sun, Yi~Tay, Jialong Han, and Chenliang Li. 2018.
\newblock Pair-linking for collective entity disambiguation: Two could be
  better than all.
\newblock \emph{IEEE Transactions on Knowledge and Data Engineering}.

\bibitem[{Raiman and Raiman(2018)}]{raiman2018deeptype}
Jonathan Raiman and Olivier Raiman. 2018.
\newblock Deeptype: Multilingual entity linking by neural type system
  evolution.
\newblock \emph{arXiv preprint arXiv:1802.01021}.

\bibitem[{Ratinov et~al.(2011)Ratinov, Roth, Downey, and
  Anderson}]{ratinov2011local}
Lev Ratinov, Dan Roth, Doug Downey, and Mike Anderson. 2011.
\newblock Local and global algorithms for disambiguation to wikipedia.
\newblock In \emph{Proceedings of the 49th Annual Meeting of the Association
  for Computational Linguistics: Human Language Technologies-Volume 1}, pages
  1375--1384. Association for Computational Linguistics.

\bibitem[{Shen et~al.(2015)Shen, Wang, and Han}]{shen2015entity}
Wei Shen, Jianyong Wang, and Jiawei Han. 2015.
\newblock Entity linking with a knowledge base: Issues, techniques, and
  solutions.
\newblock \emph{IEEE Transactions on Knowledge and Data Engineering},
  27(2):443--460.

\bibitem[{Shen et~al.(2012)Shen, Wang, Luo, and Wang}]{shen2012linden}
Wei Shen, Jianyong Wang, Ping Luo, and Min Wang. 2012.
\newblock Linden: linking named entities with knowledge base via semantic
  knowledge.
\newblock In \emph{Proceedings of the 21st international conference on World
  Wide Web}, pages 449--458. ACM.

\bibitem[{Srivastava et~al.(2014)Srivastava, Hinton, Krizhevsky, Sutskever, and
  Salakhutdinov}]{srivastava2014dropout}
Nitish Srivastava, Geoffrey Hinton, Alex Krizhevsky, Ilya Sutskever, and Ruslan
  Salakhutdinov. 2014.
\newblock Dropout: a simple way to prevent neural networks from overfitting.
\newblock \emph{The Journal of Machine Learning Research}, 15(1):1929--1958.

\bibitem[{Sutton and Barto(1998)}]{sutton1998reinforcement}
Richard~S Sutton and Andrew~G Barto. 1998.
\newblock \emph{Reinforcement learning: An introduction}, volume~1.
\newblock MIT press Cambridge.

\bibitem[{Xu and Barbosa(2018)}]{xu2018neural}
Peng Xu and Denilson Barbosa. 2018.
\newblock Neural fine-grained entity type classification with hierarchy-aware
  loss.
\newblock \emph{north american chapter of the association for computational
  linguistics}, 1:16--25.

\bibitem[{Xue et~al.(2019)Xue, Cai, Su, Song, Ge, Liu, and
  Wang}]{xue2019neural}
Mengge Xue, Weiming Cai, Jinsong Su, Linfeng Song, Yubin Ge, Yubao Liu, and Bin
  Wang. 2019.
\newblock Neural collective entity linking based on recurrent random walk
  network learning.
\newblock \emph{arXiv preprint arXiv:1906.09320}.

\bibitem[{Yamada et~al.(2016)Yamada, Shindo, Takeda, and
  Takefuji}]{yamada2016joint}
Ikuya Yamada, Hiroyuki Shindo, Hideaki Takeda, and Yoshiyasu Takefuji. 2016.
\newblock Joint learning of the embedding of words and entities for named
  entity disambiguation.
\newblock \emph{arXiv preprint arXiv:1601.01343}.

\bibitem[{Yang et~al.(2018)Yang, Irsoy, and Rahman}]{yang2018collective}
Yi~Yang, Ozan Irsoy, and Kazi~Shefaet Rahman. 2018.
\newblock Collective entity disambiguation with structured gradient tree
  boosting.
\newblock \emph{arXiv preprint arXiv:1802.10229}.

\end{thebibliography}
\bibliographystyle{acl_natbib}

\end{document}